\documentclass[runningheads]{llncs}

\usepackage{graphicx}

\begin{document}

\pagenumbering{gobble}

\title{Text Similarity Using Word Embeddings to Classify Misinformation}

\author{Caio Sacramento de Britto Almeida\inst{1,2}\and
Débora Abdalla Santos\inst{1}}
\authorrunning{C. S. B. Almeida, D. A. Santos}

\institute{
Computer Science Department, Federal University of Bahia, Brazil
\email{\{caiosba,abdalla\}@dcc.ufba.br}\\
\url{http://www.dcc.ufba.br} \and
Meedan, San Francisco, USA\\
\email{caio@meedan.com}\\
\url{https://meedan.com}}

\maketitle

\begin{abstract}
Fake news is a growing problem in the last years, especially during elections. It’s hard work to identify what is true and what is false among all the user generated content that circulates every day. Technology can help with that work and optimize the fact-checking process. In this work, we address the challenge of finding similar content in order to be able to suggest to a fact-checker articles that could have been verified before and thus avoid that the same information is verified more than once. This is especially important in collaborative approaches to fact-checking where members of large teams will not know what content others have already fact-checked.

\end{abstract}

{\let\thefootnote\relax\footnotetext{Copyright \textcopyright\ 2020 for this paper by its authors. Use permitted under Creative Commons License Attribution 4.0 International (CC BY 4.0). DHandNLP, 2 March 2020, Evora, Portugal. }}

\section{Introduction}

Fake news have always been around, but they are turning into a growing problem and more evident on the last years with the popularization of the Internet as a source of news, a role that used to be played by traditional media like television, radio, magazines and newspapers. From a theoretical point of view \cite{ling}, fake news must comply to a few characteristics: should be published and shared on the Internet; should be created with fake content and without support; is used to manipulate.

Disinformation (that is, purposely false information) has been playing a fundamental role on recent democratic electoral processes. Social media have an evident merit to allow debates and to amplify voices in a space that allows great repercussion. Many studies show \cite{enli} how Twitter, Facebook and other platforms, became important instruments of democracy as they allow exchanges and stimulate discussions. However, in the same way it happens on the public debate outside the virtual world, social media are used to disseminate fake information. Automated accounts that make it easier to send a massive amount of messages have turned into a potential tool to manipulate debates on social networks, especially in moments of political relevance.

in this way, online platforms allow old defamation strategies and public debate manipulation, now at a larger scale \cite{ruediger}. Both bots and humans have an important role in creating and spreading fake news in electoral contexts---sometimes on purpose, other times by mistake. Due to the nature of human psychology, people tend to believe things that support their existing beliefs, a process called ``confirmation bias'', defined as the possibility to remember, interpret or research in a way that confirms an initial belief or hypothesis \cite{plous}.

Attention begets more attention on social media \cite{hale2016digital}. Fake news on Twitter, for example, has 70\% more chances of being accessed than true information \cite{langin}, often due to shocking titles or sensationalism. 

\section{Automatic recognition of fake news}

Identify fake news automatically is not a trivial task. One approach is based on linguistics and try to identify text properties, like writing style and content, that can help on differentiating false articles from true articles. An assumption for this approach is that linguistic behaviors, such as punctuation, types of words and emotional charge are unconscious and thus are out of the author’s control, and those things could reveal important insights about the nature of the text.
 
Studies based on that approach reach an accuracy near 76\% when compared to human performance \cite{perez}. Although promising, future effort should not be limited by that and could also include other information, for example: number of inbound and outbound links, number of comments, visual analysis of the page where the content is at, among other computational techniques for fact verification \cite{thorne}.

Even this way, since it is a work of high responsibility, a completely automated classification of news can be very risky, and the reputation of a media organization can be affected by that. So, the purpose of this work is a hybrid approach that creates a human-in-the-loop system.

\section{Purpose}

The purpose of this work is to identify articles that are similar to articles that were previously classified by a fact-checking agent as true or false, and in this way, optimize the process of verification. The idea to achieve that is to implement a plug-in for the open source software for collaborative fact-checking Check \cite{meedan}, already used on collaborative fact-checking projects around the world. Check has been used, for example, during the US elections in 2016, France elections in 2017, Mexico elections in 2018 and Indian elections in 2019 as well as other verification projects not related to elections. Such a plug-in can optimize the fact-checking process by suggesting similar articles that were already classified in the past, avoiding that the same, or substantially similar, content is fact-checked more than once. Technically, the idea is to use artificial neural networks to identify similar articles based on their vector representations. The plugin was implemented as a Check Bot, so every time a new piece of content is created on Check, the plugin would look for similar items that had already been fact-checked.

\section{Artificial Neural Network}

Word2Vec \cite{mikolov} is the neural network that is going to be used on this work. It contains two layers and processes text. The input is a text corpus and the output is a set of vectors: characteristic vectors for words from that corpus. Its applications go beyond text analysis. It can be used for gens, code, music playlists, social network graphs, and other verbal or symbolic series where a pattern can be recognized.

An example given by the authors of Word2Vec is: the vector that represents the token “Madrid”, subtracted from the vector that represents “Spain” and summed to the vector “France” will be very close to the vector obtained for “Paris”. Or, as an equation:

\begin{equation}
vec(Madrid) - vec(Spain) + vec(France) = vec(Paris)
\end{equation}

\section{Results}

Given the great results that Word2Vec can achieve on identifying text patterns \cite{goldberg}, it was the choice for this work. 
%Ideally, the scope of this work could be extended and the input corpus for the neural network would be a large corpus of news classified in Portuguese, and this way, it would be feasible to get a neural network that could determine with reasonable accuracy the probability of any article to be true or false. However, such database does not exist yet and it would be a lot of work to build it, since a lot of news would need to be collected and classified. So, in this work, 
We use the pretrained Word2Vec model, which contains 3 million vectors of 300 dimensions \cite{mihaltz} trained on over 100 billion words from Google News.
There are several promising avenues for extending this work to Portuguese (and other languages), but the datasets currently available in Portuguese are not as extensive as Google News or as well tested; so, in this paper the plug-in is used to identify the similarity between an input article in English and pre-classified articles in English.

The flow works as the following: when new information to be verified is inserted in Check’s database, the plug-in developed in this work takes action. It calculates the vectors of this input text,  using Word2Vec, and stores the vectors in an ElasticSearch database, which is a distributed search service, open source, scalable, and has good search performance \cite{banon}. Another useful feature is that ElasticSearch can be extended through plug-ins, including plug-ins for searching criteria. For this work, it was necessary to implement a search plug-in for ElasticSearch that could calculate the similarity between the input text (represented as vectors) and each stored text (also represented as vectors) using cosine distance. The interface between Check and ElasticSearch is filled by Alegre, an API part of the Check suite which is responsible for text and image processing, for example, similarity, classification, glossary and language identification.

\begin{figure}
\includegraphics[width=\textwidth]{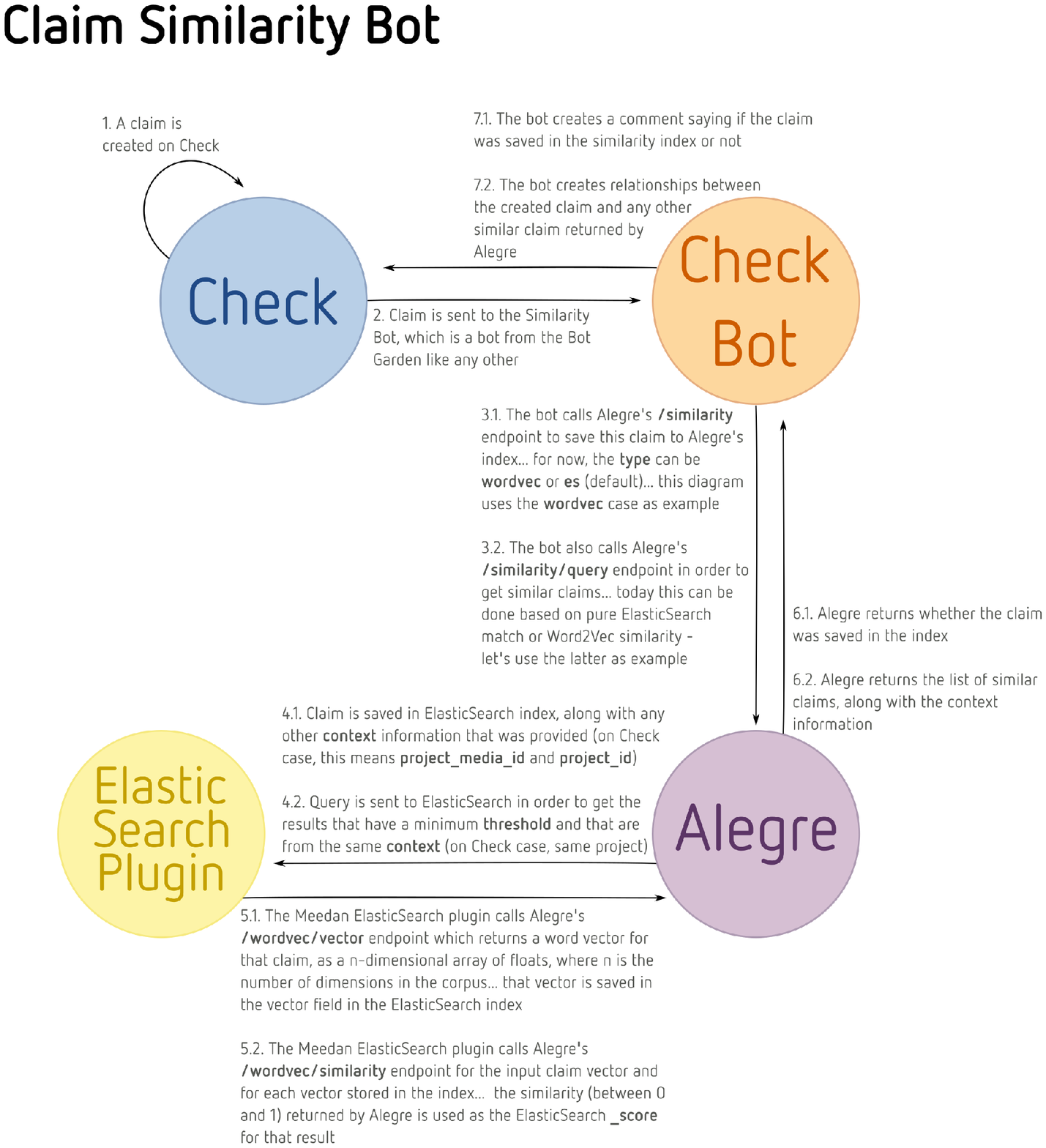}
\caption{Workflow}
\label{fig1}
\end{figure}

Therefore, searching for similar texts returns the vectors with smaller cosine distances for the input vectors. The next step of the plug-in is that it uses Check’s API in order to suggest to journalists using Check  similar articles that were classified before and that are similar to the article that the user is currently verifying. This way, the user can decide to relate them or not---and thus, avoid fact-checking the same content multiple times. This workflow is represented in Fig.~\ref{fig1}.

The plugin was integrated into the software Check and is available on GitHub \cite{check}.

\section{Conclusions and future work}

In this paper we developed an open-source architecture to use ElasticSearch to efficiently search language volumes of text items by their vector representations in near real-time and integrated this into an open-source fact-checking software tool. We used Word2Vec trained on English-language text, but the architecture we developed is not specific to these choices and can easily be adapted to other vector-representation models and languages.

First, we would like to use a corpus with Portuguese text \cite{monteiro}, so this solution could be more useful for media organizations in Brazil. Second, the notion of “similar” is too broad. Texts can be similar but completely contradictory. In this sense, it would be very useful to determine if a given text supports or refutes another text. An approach for that could be through stance detection \cite{ghanem}, but it requires more research. Moreover, the choice for Word2Vec led to promising results, but other recent advances in this domain suggest that transformer models such as BERT, SBERT and XML-RoBERTa representations are able to improve the performance of NLP tasks so those options should also be evaluated \cite{devlin,reimers,conneau}. Although suggesting similar texts can optimize  fact-checking work, much more could be done if more specific input corpora could be built. Finally, we should be able to evaluate how much this approach helps the verification work, for example, by counting how many content items did not need to be verified again because similar, previously-verified items were correctly suggested by this tool.

\end{document}